\documentclass[letterpaper, 10 pt, conference]{ieeeconf}  

\IEEEoverridecommandlockouts                              

\usepackage{graphicx} 
\usepackage{amsmath} 
\usepackage{amssymb}  
\usepackage{amsfonts}
\usepackage{color}

\usepackage{cite}

\usepackage{epstopdf}
\usepackage{wrapfig}
\usepackage{adjustbox}
\usepackage{booktabs, makecell}

\usepackage{multirow}
\usepackage{times}
\usepackage{textcomp}
\DeclareMathAlphabet{\mathcal}{OMS}{cmsy}{m}{n}
\usepackage[normalem]{ulem}
\usepackage{hhline}
\usepackage{diagbox}
\usepackage{balance}
\usepackage{mathtools}
\usepackage{algorithm}
\usepackage{algpseudocode}
\usepackage{tikz}
\usepackage{hyperref}
\usepackage{lipsum}

\usepackage[utf8]{inputenc}
\usepackage[english]{babel}

\usepackage{amsthm}

\newtheorem{theorem}{Theorem}

\newcommand\widebar[1]{\mathop{\overline{#1}}}
\newcommand{\defeq}{\vcentcolon=}

\newcommand{\rrefdir}[1]{{\widebar{\varphi}}^*_{#1}}
\newcommand{\prefdir}[1]{{\widebar{\varphi}}'_{#1}}
\newcommand{\nrefdir}[1]{{\widebar{\varphi}}_{#1}}

\newcommand\copyrighttext{%
  \footnotesize \textcopyright 2019 IEEE.  Personal use of this material is permitted.  Permission from IEEE must be obtained for all other uses, in any current or future media, including reprinting/republishing this material for advertising or promotional purposes, creating new collective works, for resale or redistribution to servers or lists, or reuse of any copyrighted component of this work in other works.
  }
\newcommand\copyrightnotice{%
\begin{tikzpicture}[remember picture,overlay]
\node[anchor=south,yshift=10pt] at (current page.south) {\fbox{\parbox{\dimexpr\textwidth-\fboxsep-\fboxrule\relax}{\copyrighttext}}};
\end{tikzpicture}%
}

\title{\LARGE \bf
Active Learning for Risk-Sensitive Inverse Reinforcement Learning
}

\author{Rui Chen, Wenshuo Wang, {\it Member, IEEE}, Zirui Zhao and Ding Zhao
\thanks{Corresponding Authors: Wenshuo Wang and Ding Zhao.}
\thanks{R. Chen is with the Department of Mechanical Engineering, Carnegie Mellon University, Pittsburgh, PA, 15213, and also with the Department of Electrical Engineering and Computer Science, University of Michigan, Ann Arbor, MI, 48109
        {\tt\small richen@umich.edu}}%
\thanks{W. Wang, and D. Zhao are with Carnegie Mellon University, Pittsburgh, PA, 15213
        {\tt\small \{wenshuow, dingzhao\}@andrew.cmu.edu}}%
\thanks{Z. Zhao is with the School of Electronic and Information Engineering, Xi'an Jiaotong University,
        Xi'an, Shaanxi 710049, China.
        {\tt\small ryan\_zzr@outlook.com}}%
}

\begin{document}

\maketitle
\copyrightnotice
\thispagestyle{empty}
\pagestyle{empty}

\begin{abstract}
One typical assumption in inverse reinforcement learning (IRL) is that human experts act to optimize the expected utility of a stochastic cost with a fixed distribution. This assumption deviates from actual human behaviors under ambiguity. Risk-sensitive inverse reinforcement learning (RS-IRL) bridges such gap by assuming that humans act according to a random cost with respect to a set of subjectively distorted distributions instead of a fixed one. Such assumption provides the additional flexibility to model human's risk preferences, represented by a risk envelope, in safe-critical tasks. However, like other learning from demonstration techniques, RS-IRL could also suffer inefficient learning due to redundant demonstrations. Inspired by the concept of active learning, this research derives a probabilistic disturbance sampling scheme to enable an RS-IRL agent to query expert support that is likely to expose unrevealed boundaries of the expert's risk envelope. Experimental results confirm that our approach accelerates the convergence of RS-IRL algorithms with lower variance while still guaranteeing unbiased convergence.

\end{abstract}

\section{Introduction}
Inverse reinforcement learning (IRL) provides a novel framework for recovering cost functions utilized in human decision making \cite{ng_algorithms_2000, abbeel_apprenticeship_2004, englert_inverse_2017, levine_continuous_2012, ramachandran_bayesian_2007, ziebart_maximum_2008}. The original IRL algorithms \cite{ng_algorithms_2000, abbeel_apprenticeship_2004} are formed as linear programming constrained by optimality conditions \cite{zhifei_survey_2012}. Some major IRL variants are also proposed, e.g., Maximum margin planning (MMP) \cite{ratliff_maximum_2006, ratliff_learning_2009}, Bayeisn IRL \cite{ramachandran_bayesian_2007}, and Maximum entropy (MaxEnt) IRL \cite{ziebart_maximum_2008}. More recent advancements in IRL include the guided cost learning algorithm \cite{finn_guided_2016} which combines MaxEnt IRL and deep learning techniques. The flexibility of IRL framework has prompted its application to a variety of tasks such as autonomous helicopter aerobatics \cite{abbeel_autonomous_2010}
and robot locomotion \cite{park_inverse_2013}. Most IRL algorithms adopt expected utility theory (EUT) \cite{von_neumann_theory_1944}, which assumes that human experts act in order to optimize the statistical expectation of a utility function of stochastic costs. However, EUT lacks its ability to model human behavior in risky scenarios and does not distinguish the scenarios where outcome probabilities are known to human and ones where they are not, as discussed by Majumdar~\textit{et al.}~\cite{majumdar_risk-sensitive_2017} and Singh~\textit{et al.}~\cite{singh_risk-sensitive_2018}. Thus, robots reasoning based on EUT could make biased assumptions about human behavior and thus lead to catastrophic consequences under safe-critical settings such as autonomous driving.

The inconsistency between human behavior and EUT under unknown outcome probabilities can be illustrated using the \textit{Ellsberg paradox} \cite{ellsberg_risk_1961}. In a hypothetical experiment, certain amount of red and black balls exist in two independent urns, from one of which a ball will be drawn randomly. Human participants are provided with following pieces of information: (1) Urn I contains 100 red and black balls but in an unknown ratio; (2) Urn II contains 50 red balls and 50 black balls; (3) \$100 reward for a red ball and \$0 for a black one. People are then asked to choose one of the urns and randomly draw a ball from it. It turns out that people hold a dominating preference towards Urn II over Urn I. However, in another setting where the \$100 reward will be granted to a black ball (no reward for red), people still prefer Urn II. Applying EUT with a utility function for two outcomes (\$100 and \$0), we deduce that people assess a lower probability of drawing red balls from Urn I in the first setting (\$100 for red), while in the second setting (\$100 for black), they assess a higher probability of drawing red balls from Urn I. The contradictory behavior suggests that EUT may fail to explain scenarios where outcome probabilities are unknown, i.e., \textit{ambiguous}, to humans.


One way to interpret the above paradox is that apart from being risk averse, humans are also \textit{ambiguity averse} \cite{gilboa2016ambiguity}. Since EUT assumes that a fixed outcome distribution is available to and utilized by humans, it lacks the flexibility to capture the human prospect when ambiguity is present. To resolve the above limitation, Majumdar~\textit{et al.}~\cite{majumdar_risk-sensitive_2017} and Singh~\textit{et al.}~\cite{singh_risk-sensitive_2018} propose a risk-sensitive inverse reinforcement learning (RS-IRL) framework, which models humans as optimizing costs with respect to a subclass of coherent risk measures (CRM). 
In RS-IRL setting, when the outcome probabilities are unknown, humans act to minimize (maximize) the worst-case expected cost (reward) with respect to a set of subjectively distorted outcome probabilities, or \textit{risk envelope}. The goal of RS-IRL is then to infer human's risk envelope from demonstrations. The introduction of CRM provides significant flexibility in capturing individual risk preferences and thus makes RS-IRL suitable for modeling human decision making within ambiguity. Notably, the Ellsberg paradox can be resolved by assuming the maximum ambiguity: the human risk envelope spans the entire probability simplex. In other words, in human's perspective, the probability of drawing red or black balls from Urn I could be any value between 0 and 1 depending on reward settings. The worst-case expected reward of choosing Urn I is then always zero since the human could always assess zero probability of drawing rewarding balls. On the other hand, choosing Urn II leads to a deterministic \$50 expected reward and is constantly preferred by human participants.

Majumdar~\textit{et al.}~\cite{majumdar_risk-sensitive_2017} first solved the RS-IRL problem via linear programming (LP) based algorithms that recursively refine the estimate of risk envelope via half-plane constraints given expert demonstrations. Singh~\textit{et al.}~\cite{singh_risk-sensitive_2018} generalized the work to sequential disturbances setting and formally proved that with sufficient optimal demonstrations, the single-step RS-IRL would exactly replicate the expert's decision making. Although the above methods succeed in estimating human risk envelope in realistic tasks such as car-following, the result can be under-refined due to redundant demonstrations. 
For instance, the inferred risk envelope for a participant in the multi-step driving game \cite{majumdar_risk-sensitive_2017} spans almost the entire probability simplex. In a similar driving simulation with multi-stage planning, the inferred risk envelope shows large ambiguity over one of the disturbances and yields preemptive predictions comparing to actual observations \cite{singh_risk-sensitive_2018}. Additional training data would indeed improve this situation but is not always available. 
For instance, an industrial robot collaborating with different workers should be able to infer each worker's risk preferences within the limited task execution period in a ``one-shot'' manner. We further notice that the convergence of original RS-IRL algorithm could suffer non-trivial variance, which also indicates slow convergence in practice.

This paper incorporates the concept of active learning to address the above challenges under single-step and multi-step cases \cite{majumdar_risk-sensitive_2017}. Active learning enables the agent to query demonstrations from the expert on states where the policy has not been well observed \cite{lopes_active_2009, odom_active_2016} or to decide when to ask for expert help depending on relative utilities versus acting on its own \cite{cohn_selecting_2010}. 
In the original RS-IRL setting, the disturbance is sampled according to a fixed distribution. However, robot and its actions normally are the major sources of disturbance to humans. Therefore, the robot should be able to influence the disturbance sampling via its own actions. This provides the robot with a real handle to facilitate future learning based on acquired knowledge. 
This work derives an adaptive disturbance sampling scheme to enable an RS-IRL learner to query expert's demonstrations that are likely to expose unrevealed risk envelope boundaries based on past learning experiences.

The rest of paper is structured as follows. Section \ref{sec:problem} briefly introduces the RS-IRL problem. Section \ref{sec:single} explains the single-step RS-IRL algorithm with active learning. Section \ref{sec:multi} introduces a generalization to multi-step RS-IRL. Section \ref{sec:experiments} shows the experiments and results. Section \ref{sec:conclusion} concludes our work with future works.


\section{Problem Formulation}\label{sec:problem}

\subsection{Dynamics and Expert Model}
Consider a discrete-time dynamical system with state ${x_k\in\mathbb{R}^n}$, action $u_k\in\mathbb{R}^m$, and disturbance $w_k\in\mathcal{W}$ as
\begin{equation}
\label{eq:dynamics}
    x_{k+1} = f(x_k, u_k, w_k)
\end{equation}
The action $u\in\mathcal{U}$ is bounded component-wise, i.e., ${u^-(j)\leq u(j)\leq u^+(j),~\forall j\in\{1\dots m\}}$. The disturbance set $\mathcal{W}\defeq\{w^{[1]},w^{[2]},\dots,w^{[L]}\}$ is finite with probability mass function (pmf) $p\defeq[p(1), p(2),\dots,p(L)]$. The $w_k$ is sampled differently under single-step and multi-step settings and will be explained in Section~\ref{sec:single} and Section~\ref{sec:multi}. The human expert is assumed to know decision-making dynamics \eqref{eq:dynamics} and disturbance realizations $\mathcal{W}$, but not the pmf $p$.

Let $C(x_k, u_k, w_k)$ denote the cost function of a state-action pair $(x_k, u_k)$ when $w_k$ is realized. Let $Z$ denote the cumulative cost calculated by the human expert over a finite planning horizon. Under stochastic disturbances, $Z$ is a random variable that has a realization for each trajectory $\{(x_k, u_k, w_k)\}$. The human expert is then assumed to be optimizing a CRM defined on $Z$ according to the following representation theorem.


\begin{theorem}[Representation theorem for CRMs \cite{artzner_coherent_1999}]
 Suppose $\Omega$ is a countable sample space with cardinality $|\Omega|$, $\mathcal{F}=2^\Omega$ is the $\sigma$-field, and $\mathcal{Z}$ is the space of random variables defined on $\Omega$. A risk measure $\rho:\mathcal{Z}\rightarrow\mathbb{R}$ is coherent if and only if there exists a family $\mathcal{P}$ of probability measures such that for any $Z\in\mathcal{Z}$:
 \begin{equation}
 \label{def:crm}
 \rho(Z)=\underset{q\in\mathcal{P}}{\mathrm{max}~}\mathbb{E}_q[Z]=\underset{q\in\mathcal{P}}{\mathrm{max}~}\sum_{i=1}^{|\Omega|}q(i)Z(i)
 \end{equation}
\end{theorem}

Conceptually, a CRM $\rho(\cdot)$ is calculating the worst-case expected value of a random cost $Z$ with respect to a set of probability measures $\mathcal{P}$ defined on the same sample space $\Omega$. Therefore, a human expert minimizing a CRM is ambiguous about the true disturbance distribution. In other words, any probability measure $q\in\mathcal{P}$ can be regarded as the true distribution $p$ distorted by a subjective distortion $\zeta$, i.e., $q(i) = p(i)\cdot\zeta(i),~\forall i\in{1,2\dots,|\Omega|}$ with $\sum_i q(i)=1$. Thus, estimating the set $\mathcal{P}$ is equivalent to inferring the human expert's risk preferences, which forms the basis of the RS-IRL methodology.
In RS-IRL setting, $\mathcal{P}$ is taken to be a polytopic subset of the probability simplex $\Delta^{|\Omega|}$ subject to $d$ half-space constraints:
\begin{equation}
    \mathcal{P} = \{q\in\Delta^{|\Omega|}~|~A_{ineq}q\leq b_{ineq}\}
\end{equation}
with $A_{ineq}\in\mathbb{R}^{d\times|\Omega|}$ and $b_{ineq}\in\mathbb{R}^d$. The goal of RS-IRL problem is to infer the risk preferences of an expert from demonstrations $\{(x^*_k,u^*_k)\}_{k=1}^{D}$ by estimating $\mathcal{P}$, which is referred to as the \textit{risk envelope} hereafter.

\subsection{Active Learning for RS-IRL}
For observed expert demonstration $(x^*_k,u^*_k)$ on each decision making period, the original RS-IRL algorithms proceed by computing and applying a half-plane constraint $\mathcal{H}_{(x^*_k,u^*_k)}$ on $\mathcal{P}$ (see Section \ref{sec:single}). Although $\mathcal{P}$ can be efficiently approximated by sequentially solving LP problems, our experiments (see Section \ref{sec:single}) show that given certain disturbance sequences, the convergence is slow due to redundant constraints (see, e.g., Fig.~\ref{fig:std_redundant}). The goal of this paper is an adaptive sampling scheme to generate disturbances $\{w_k\}$ for querying expert demonstrations that are likely to generate non-redundant constraints and expose unrevealed risk envelope boundaries.

\section{Single-Step Decision Making}
\label{sec:single}

\subsection{Single-Step RS-IRL Model}
We first introduce the single-step decision model \cite{majumdar_risk-sensitive_2017} with the known cost function. At each time step $k$, the expert generates an action $u_k$ to minimize a CRM of a random cost $Z$. $Z$ is a random variable on the discrete probability space $(\mathcal{W}, 2^{\mathcal{W}}, p)$, where $p$ refers to the true probability measure that is unknown to human. The $j^{th}$ realization of $Z$ is calculated via a known random cost function $g$ for current state-action pair $(x_k, u_k)$ with disturbance realization $w^{[j]}$.
The expert is then selecting an action that optimizes the following problem:
\begin{align}
    \underset{u_k\in\mathcal{U}}{\mathrm{min}}~\rho(Z) &= \underset{u_k\in\mathcal{U}}{\mathrm{min}}~\underset{q\in\mathcal{P}}{\mathrm{max}}~\mathbb{E}_q[Z] = \underset{u_k\in\mathcal{U}}{\mathrm{min}}~\underset{q\in\mathcal{P}}{\mathrm{max}}~g(x_k, u_k)^Tq\label{optim:single_minmax}
\end{align}
where $\mathcal{P}$ is a polytopic subset of the probability simplex $\Delta^{|\Omega|}$ ($|\Omega|=L$). Since the objective is linear in $q$, given $u_k$, the optimal value for the maximization part must be achieved at a vertex of the polytope $\mathcal{P}$. Let $\mathcal{V}(\mathcal{P})=\{v_i\}$ denote the set of vertices of $\mathcal{P}$, problem \eqref{optim:single_minmax} can be rewritten as:
\begin{subequations}\label{optim:single_alt}
\begin{align}
    \tau^*=\underset{u_k\in\mathcal{U}, \tau}{\mathrm{min}}~~~&\tau\tag{\ref{optim:single_alt}}\\
    \mathrm{s.t.}~~~&\tau\geq g(x_k, u_k)^Tv_i,~\forall v_i\in\mathcal{V}(\mathcal{P})\label{optim:single_alt_constraint}
\end{align}
\end{subequations}
If the cost vector $g(x_k, u_k)\in\mathbb{R}^L$ is convex in action $u$, the optimization problem \eqref{optim:single_alt} is also convex.

\begin{figure}[t]
    \centering
    	\includegraphics[width=0.9\columnwidth]{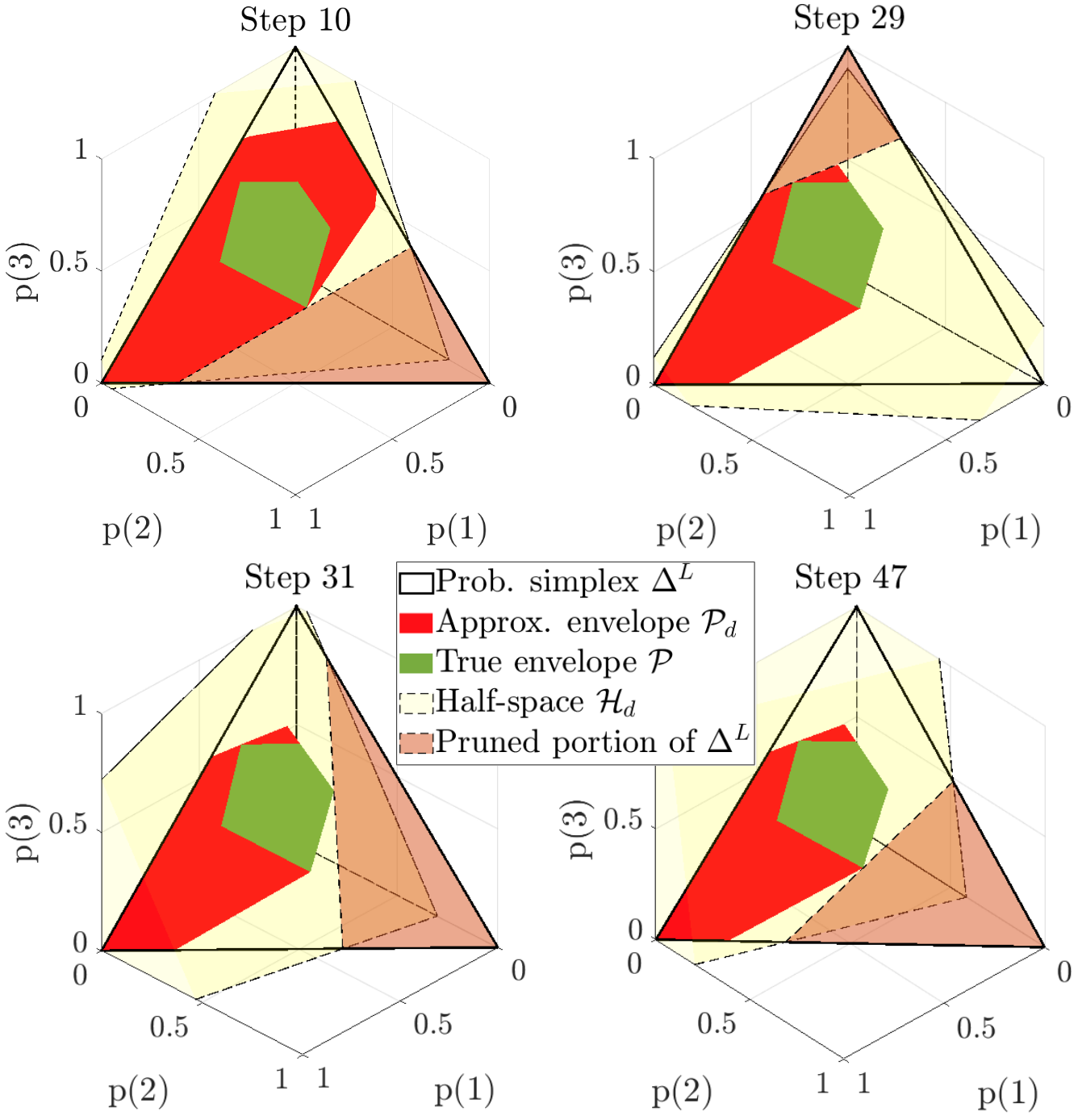}
    \caption{Example of risk envelope refinement using original single-step RS-IRL model with $L=3$. At steps 10 and 29, the half-space constraint $\mathcal{H}_{d}$ prunes a portion of the probability simplex and successfully refine the envelope approximation. In steps 31 and 47, however, the demonstrations yield redundant constraints that fail to refine the outer envelope approximation. The yellow-shaded volume indicates the feasible portion of the first octant that satisfies the constraint implied by $\mathcal{H}_d$.}
    \label{fig:std_redundant}
\end{figure}

Given an optimal state-action pair $(x^*, u^*)$, the RS-IRL algorithm proceeds by examining the Karush-Kuhn-Tucker (KKT) conditions for problem \eqref{optim:single_alt}, which are necessary for optimality and also sufficient in convex problems. The KKT conditions will constrain the constraints in problem \eqref{optim:single_alt} (i.e., $\mathcal{V}(\mathcal{P})$) to ensure that the given demonstration $(x^*, u^*)$ is indeed the optimal solution to \eqref{optim:single_alt}. The number of vertices of the true envelope $\mathcal{P}$ is assumed to be unknown. Define $\mathcal{J}^+\defeq\{j\in\{1,\dots,m\}~|~u^*(j)=u^+(j)\}$ and $\mathcal{J}^-\defeq\{j\in\{1,\dots,m\}~|~u^*(j)=u^-(j)\}$ as the indices of saturated action components, 
thus, the expert risk envelope is constrained according to the following theorem:
\begin{theorem}[KKT-based inference \cite{majumdar_risk-sensitive_2017, singh_risk-sensitive_2018}]
\label{thm:single_kkt}
Given an optimal state action pair $<x^*, u^*>$, define the half-space:
\begin{equation}
\label{def:single_half_space}
    \mathcal{H}_{(x^*, u^*)}\defeq\{v\in\mathbb{R}^L~|~\tau'\geq g(x^*, u^*)^Tv\}
\end{equation}
The risk envelope $\mathcal{P}$ satisfies
\begin{equation}
\label{constr:H}
\mathcal{P}\subset(\mathcal{H}_{(x^*, u^*)}\cap\Delta^L)    
\end{equation}
where $\tau'$ refers to the optimal objective of the following optimization problem:
\begin{align}
    \max_{\substack{v\in\Delta^L\\\sigma_+, \sigma_-\geq 0}} ~&g(x^*, u^*)^Tv \label{optim:single_kkt}\\
    s.t.~~&\nabla_{u(j)}g(x,u)^Tv\Big\vert_{x^*, u^*} + \sigma_+(j) = 0,~\forall j\in\mathcal{J}^+\nonumber\\
        ~~&\nabla_{u(j)}g(x,u)^Tv\Big\vert_{x^*, u^*} - \sigma_-(j) = 0,~\forall j\in\mathcal{J}^-\nonumber\\
        ~~&\nabla_{u(j)}g(x,u)^Tv\Big\vert_{x^*, u^*} = 0,~\forall j\not\in\mathcal{J}^+,j\not\in\mathcal{J}^-\nonumber
\end{align}
where $\sigma_+(j)$ and $\sigma_-(j)$ are multipliers for action component constraints. 
\end{theorem}
The basic idea is that according to the KKT conditions for \eqref{optim:single_alt}, the optimal objective $\tau^*$ of \eqref{optim:single_alt} is a feasible value of problem \eqref{optim:single_kkt} evaluated at $v = \bar{v} = \sum_{i\in\mathcal{I}}\lambda_iv_i$, i.e., ${g(x^*, u^*)^T\bar{v}=\tau^*}$, where $\{\lambda_i\}$ are multipliers for constraints \eqref{optim:single_alt_constraint}. $\mathcal{I}$ indicates the set of optimal vertices for problem \eqref{optim:single_alt} that satisfies ${g(x^*, u^*)^Tv_i=\tau^*,~\forall i\in\mathcal{I}}$. Therefore, solving \eqref{optim:single_kkt} would yield an optimal value $\tau'\geq\tau^*$. Each optimal state-action pair would provide a half-space constraint on $\mathcal{P}$; for any vertex $v_i$ in $\mathcal{V}(\mathcal{P})$, we have $g(x^*, u^*)^Tv_i\leq\tau^*\leq\tau'$. Let $\mathcal{P}_d$ denote the approximation of $\mathcal{P}$ after processing the first $d$ demonstrations $\{(x^*_k, u^*_k)\}_{k=1}^d$. Finally, given a new optimal state-action pair $(x^*_{d+1}, u^*_{d+1})$, the envelope approximation is updated by
\begin{equation}
    \mathcal{P}_{d+1} = \mathcal{P}_d\cap\mathcal{H}_{(x^*_{d+1}, u^*_{d+1})}.
\end{equation}
$\mathcal{P}_0$ is initialized as the probability simplex $\Delta^L$. Notably, when solving $\mathcal{H}_{(x^*_{d+1}, u^*_{d+1})}$, one can replace the constraint ${v\in\Delta^L}$ in \eqref{optim:single_kkt} by $v\in\mathcal{P}_d$ without violating Theorem \ref{thm:single_kkt} and get a tighter constraint \cite{singh_risk-sensitive_2018}.

\subsection{Adaptive Disturbance Sampling for Single-Step RS-IRL}
Upon observing an optimal demonstration $(x^*_d, u^*_d)$, a half-space constraint \eqref{constr:H} is solved to prune a portion of the probability simplex (see Fig.~\ref{fig:std_redundant} marked in orange). If this portion intersects $\mathcal{P}_{d-1}$, then $\mathcal{P}_{d}\subset\mathcal{P}_{d-1}$ which yields successful envelope refinement. However, when partial boundary of $\mathcal{P}$ has already been modeled, expert demonstrations could lead to redundant constraints. An extensive set of experiments (see Section~\ref{sec:experiments}) show that redundant constraints add non-trivial variance to the learning process. In this paper, we are particularly interested in improving the convergence rate and variance of RS-IRL algorithms by enabling the robot to actively query demonstrations that are likely to yield non-redundant constraints.

\begin{figure}
    \centering
        \includegraphics[width=0.7\columnwidth]{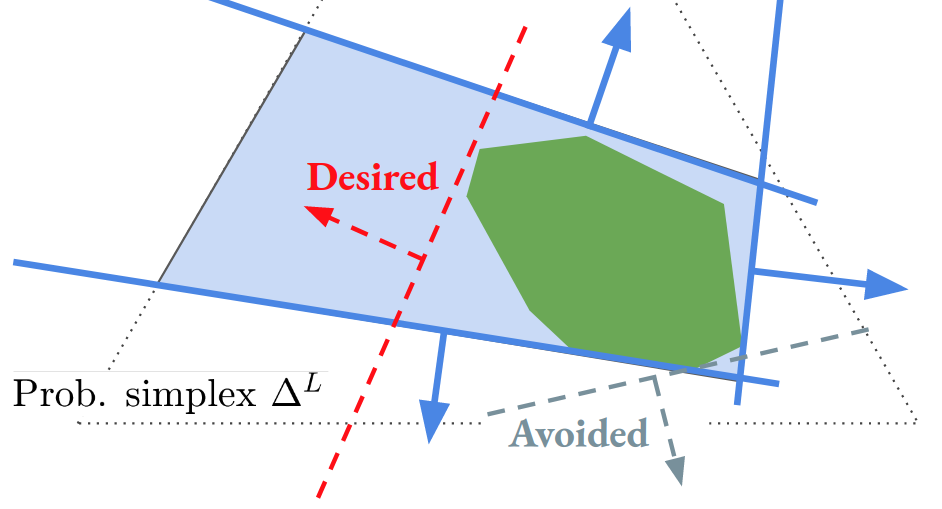}
    \caption{Given previous envelope refinement directions (blue), the learning agent prefers directions along which simplex pruning is insufficient (red) and avoids those that are already explored (grey). Blue and green polygons indicate approximated and true envelope respectively.}
    \label{fig:active_schematic}
\end{figure}

Let $\mathcal{H}'_{d+1}\defeq\{v\in\mathbb{R}^L~|~\tau'_{d+1}={g^*_{d+1}}^Tv\}$ denote the dividing hyper-plane of $\mathcal{H}_{(x^*_{d+1}, u^*_{d+1})}$, where $\tau'_{d+1}$ is the optimal value of \eqref{optim:single_kkt} solved with the ${(d+1)}^{th}$ demonstration. Then, the portion of $\Delta^L$ pruned by constraint \eqref{constr:H} is fully characterized by the segment $\mathcal{L}_{d+1}=\mathcal{H}'_{d+1}\cap\Delta^L$. The direction of $\mathcal{L}_{d+1}$ is always perpendicular to the cost vector ${g^*_{d+1}}$ and its projection on the probability simplex
\begin{equation}
\label{def:refine_dir}
    \varphi^*_{d+1}\defeq \varphi(x^*_{d+1}, u^*_{d+1})\defeq \mathrm{Proj}_{\Delta^L}(g^*_{d+1})
\end{equation}
The normalized vector $\rrefdir{d+1}$ can be determined without solving \eqref{optim:single_kkt} and is referred to as \textit{refinement direction}. The optimal value $\tau'_{d+1}$ of \eqref{optim:single_kkt} only controls how close $\mathcal{L}_{d+1}$ is to the true envelope $\mathcal{P}$ without changing its orientation. 

\begin{figure}
    \centering
        \includegraphics[width=0.9\columnwidth]{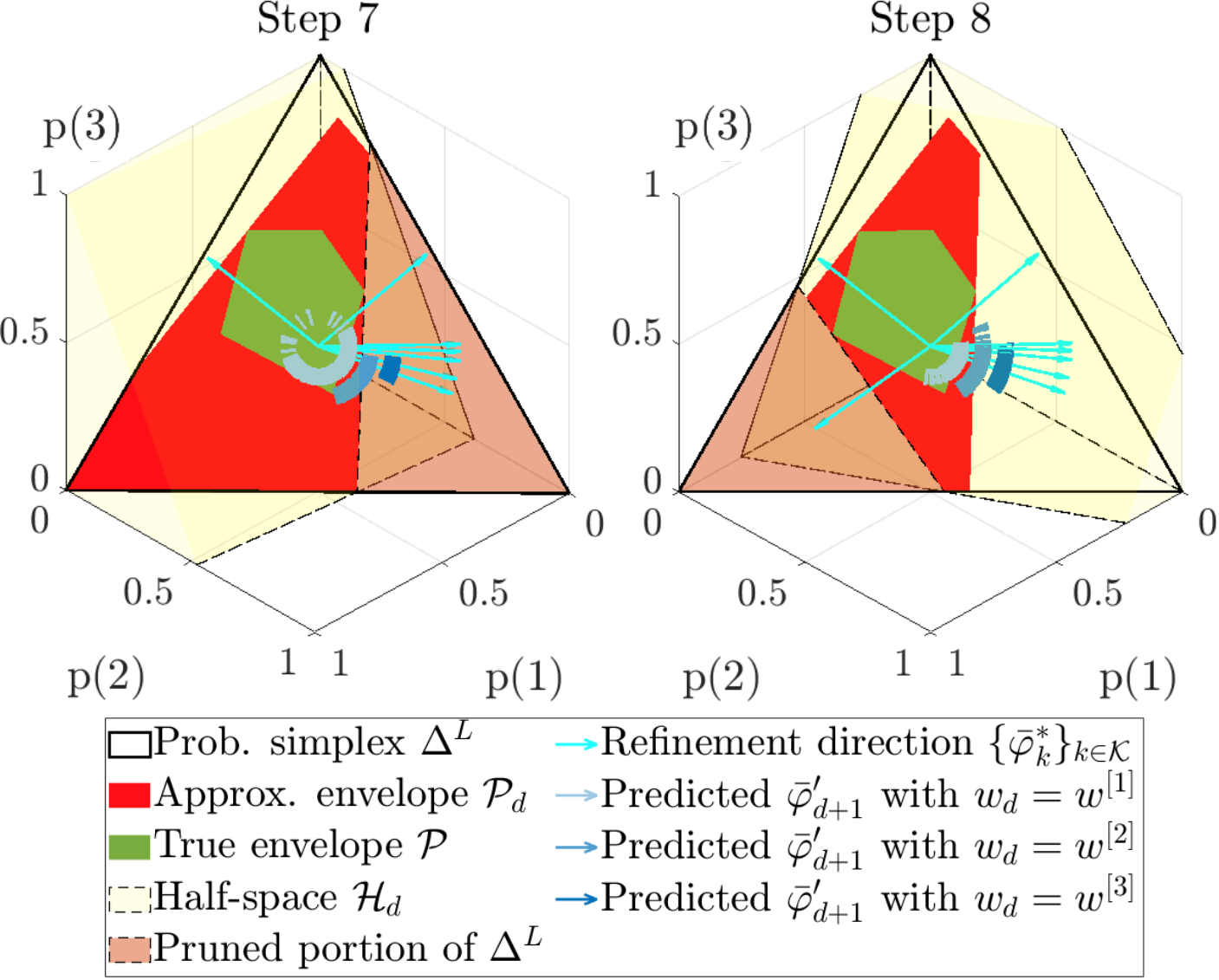}
    \caption{At step 7, 1000 predictive refinement directions $\prefdir{d+1}$ are sampled for each disturbance. The results are visualized as a circular ring where the shown segments indicate available next refinement directions. Note that for $w_d=w^{[1]}$, a large portion of the ring points toward unexplored directions. According to \eqref{eq:boltzman}, the robot samples $w_d$ with probability $p(w^{[1]}, w^{[2]}, w^{[3]})=[0.5144,~0.3127,~0.1729]$. After $w_d=w^{[1]}$ is sampled, the new demonstration yields an effective envelope refinement as expected. Note that the newly pruned area at step 8 is still not explored at step 47 in the example shown in Fig.~\ref{fig:std_redundant}.}
    \label{fig:active_exmaple}
\end{figure}

It turns out that the demonstration $(x^*_{d+1}, u^*_{d+1})$ would be redundant in terms of envelope approximation if it yields a refinement direction $\rrefdir{d+1}$ along which sufficient pruning has already been made. In Fig.~\ref{fig:std_redundant}, the demonstration produces a redundant constraint since the portion of probability simplex it prunes has already been explored in earlier operations. Determining which part of the envelope boundary has been exactly captured is generally impossible; this only happens when $\tau'$ reaches $\tau^*$ and the optimal vertices $\{v_i\}_{i\in\mathcal{I}}$ fall exactly on $\mathcal{L}$. Without knowing $\mathcal{P}$ in advance, $\tau^*$ is also unavailable. 
However, it is plausible to assume that a demonstration $(x^*_{d+1}, u^*_{d+1})$ is more likely to be informative if the corresponding refinement direction $\rrefdir{d+1}$ falls in an less explored region (i.e., dissimilar to previous $\rrefdir{k}$). Then, a disturbance more likely to lead to informative demonstrations in the next decision period should be selected by the robot. See Fig.~\ref{fig:active_schematic} for a schematic illustration.

Let $\mathcal{K}$ denote the set of time instances when expert demonstrations yield successful envelope refinements in the sense that $\forall k\in\mathcal{K}$, ${\mathcal{P}_k\subset\mathcal{P}_{k-1}}$. Then, given the trajectory $\{(x^*_k, u^*_k)\}_{k=1}^d$, we define the preference $U_{d}(j)$ for disturbance $w_d=w^{[j]}$ as the total expected dissimilarity between predicted refinement directions $\prefdir{d+1}$ and each explored directions $\{\rrefdir{k}\}_{k\in\mathcal{K}}$:
\begin{equation}
\label{def:dist_preference}
    U_d(j)=\sum_{k\sim\mathcal{K}}\mathbb{E}_{u'_{d+1}\sim\pi}\Big[-\mathcal{F}(\rrefdir{k}, \prefdir{d+1})\Big\vert_{w_d=w^{[j]}}\Big]
\end{equation}
where $\mathcal{F}(\cdot, \cdot)$ is cosine similarity and $\pi$ is the expert policy implied by the forward problem \eqref{optim:single_alt}. Conceptually, $U_d(j)$ evaluates a disturbance $w_d$ by predicting future demonstrations together with refinement directions and comparing with previous ones. Given that $\mathcal{P}$ is unknown, we replace $\pi$ with a component-wise uniform distribution, i.e., ${u'_{d+1}(j)\sim \mathrm{uniform}[u^-(j), u^+(j)]}$, and solve the expectation via sampling. Each $\prefdir{d+1}$ is predicted by projecting sampled cost vector $g(x^*_{d+1}, u'_{d+1})$ onto the probability simplex $\Delta^L$. Since the refinement direction $\prefdir{d+1}$ is jointly determined by $x^*_{d+1}$ and $u'_{d+1}$, it is bounded to a region even when only $x^*_{d+1}$ is fixed and $u'_{d+1}$ takes any available value. The more concentrated this region is, the better $U_d(j)$ is as a quality measure of disturbance $w^{[j]}$.
\begin{algorithm}
\caption{Single-Step RS-IRL with Active Learning}
\label{alg:single_active}
\begin{algorithmic}[1]
\State $\mathcal{P}_0 \gets \Delta^L$, $\mathcal{K}\gets\varnothing$
\For{$k=1$ {\bf to} $d$}
    \State Observe $(x^*_k, u^*_k)$
    \State Compute $\mathcal{H}_{(x^*_k, u^*_k)},\rrefdir{k}$ \Comment{Theorem \ref{thm:single_kkt}}
    \State $\mathcal{P}_k\gets\mathcal{P}_{k-1}\cap\mathcal{H}_{(x^*_k, u^*_k)}$
    \If {$\mathcal{P}_k\subset\mathcal{P}_{k-1}$} \State $\mathcal{K}\gets\mathcal{K}\cup k$, store $\rrefdir{k}$ \EndIf
    \State Compute $U_k(j),~\forall j\in\{1,\dots,L\}$ \Comment{\eqref{def:dist_preference}}
    \State Sample $w_k$ $\sim p(w_k = w^{[j]}) = e^{{U}_k(j)}/\sum_{j'}e^{{U}_k(j')}$
\EndFor
\State Return $\mathcal{P}_d$
\end{algorithmic}
\end{algorithm}
After receiving and processing expert demonstration at time $d$, the robot selects the disturbance $w_d=w^{[j]}$ exponentially likely with respect to $U_d(j)$:
\begin{equation}
\label{eq:boltzman}
    p(w_d=w^{[j]})\propto \mathrm{exp}(U_d(j)),~\forall j\in\{1,\dots,L\}
\end{equation}
The complete single-step RS-IRL algorithm with active disturbance sampling is summarized as \textbf{Algorithm \ref{alg:single_active}}. 
See Section \ref{sec:experiments} for an extensive experimental validation. For ease of visualization and illustration, we show examples of decision-making systems with $L=3$ disturbance realizations. 

\section{Multi-Step Planning}
\label{sec:multi}

\subsection{Multi-Step RS-IRL Model}

\begin{figure}
    \centering
    	\includegraphics[width=0.9\columnwidth]{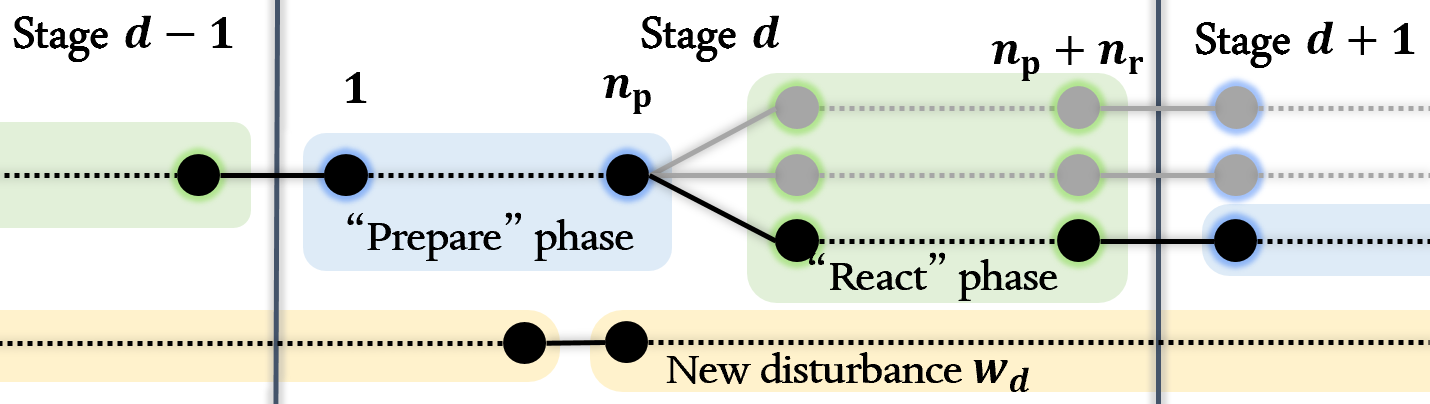}
    \caption{Scenario tree at stage $d$. A new disturbance $w_d$ is realized at the ${n_p}^{th}$ step of current stage and will be observed by the expert at the $(n_p+1)^{th}$ step. Then, the expert executes the react policy.}
    \label{fig:multistep}
\end{figure}

In this section, we introduce the extension of RS-IRL model to multi-step planning case  with unknown cost function from \cite{majumdar_risk-sensitive_2017} with some modifications. In such setting, a disturbance is sampled every $N$ steps (i.e., \textit{a stage}) and remain fixed until next sampling. Fig.~\ref{fig:multistep} shows a scenario tree for this setting. 
The planning horizon is further partitioned into two phases: ``prepare'' and ``react''. The prepare phase lasts for $n_p$ steps. At the ${n_p}^{th}$ step of each $N-$step period, a new disturbance is realized. The react phase starts from the ${(n_p+1)}^{th}$ step and lasts for $n_r$ steps. 

Given above settings, we have $w_k=w_{d-1}$ for $1\leq k\leq n_p-1$ and $w_k = w_{d}$ for $n_p\leq k\leq N$, where $k$ is the time index within each stage instead of a global time index. At the first time step of each stage $d$, the expert generates a policy $\pi_{p}:\mathcal{X}\rightarrow \mathcal{U}$ for prepare phase and a separate react policy $\pi^{[j]}_{r}:\mathcal{X}\rightarrow \mathcal{U}$ assuming each disturbance realization $j$ is observed at step $n_p+1$. Then, the optimization problem solved by the expert at step $k=1$ of stage $d$ is written as following:
\begin{align}
\label{optim:multi}
    \underset{\substack{\pi_{p},\pi^{[j]}_{r},\forall j}}{\mathrm{min}}~&C_{1:n_p-1}(x, \pi_{p}(x),w)\Big\vert_{w=w_{d-1}}+\nonumber\\
     &\rho\left(g_{n_p}(x, \pi_{p}(x)) + g_{n_p+1:N}(x, \pi_{r}(x))\right)
\end{align}
where $\rho(\cdot)$ is the CRM defined in \eqref{def:crm}. $C_{m:n}(x,u,w)$ is the cumulative cost along trajectory $\{(x_k, u_k, w_k)\}_{k=m}^n$. $g_{m:n}(x,u)\in\mathbb{R}^L$ is the cumulative random cost
\begin{equation}
g_{m:n}(x,u)(j) = \sum_{k=m}^nC(x_k, u_k, w_k)\Big\vert_{w_k=w^{[j]}},~\forall j
\end{equation}
Different from the assumption made in \cite{majumdar_risk-sensitive_2017}, the cost at step $n_p$ is stochastic. Thus, problem \eqref{optim:multi} can be written in a form similar to \eqref{optim:single_alt}:
\begin{align}
\label{optim:multi_alt}
    \underset{\substack{\tau,\pi_{p},\\\pi^{[j]}_{r},\forall j}}{\mathrm{min}}~&C_{1:n_p-1}(x, \pi_{p}(x),w)\Big\vert_{w=w_{d-1}}+\tau\\
    s.t.~&\tau\geq g(\widebar{x}, \widebar{u})^Tv_i,~\forall v_i\in \mathcal{V}(\mathcal{P})\nonumber\\
    & \pi_{p}(\cdot)\in\mathcal{U},~\pi^{[j]}_{r}(\cdot)\in\mathcal{U},~\forall j\in\{1,\dots,L\}\nonumber\\
    & x_{k+1} = f(x_k, u_k, w_k),~\forall 1\leq k\leq N\nonumber
\end{align}
where $g(\widebar{x}, \widebar{u})\in\mathbb{R}^L$ denotes the cumulative random cost of the last $(n_r+1)$ state-action pairs of a stage. We further adopt the assumption made in \cite{majumdar_risk-sensitive_2017} that the expert policies (i.e., $\pi_{p}$ and $\pi^{[j]}_{r}$) are open-loop control sequences; after executing prepare sequence ${\pi_p\defeq\{u_k\}_{k=1}^{n_p}}$, the expert selects a react sequence ${\pi^{[j]}_r\defeq\{u^{[j]}_k\}_{k=n_p+1}^{N}}$ depending on $w_d$ observed at step $n_p+1$. Leveraging the KKT conditions for problem \eqref{optim:multi_alt}, we derive an optimization problem similar to \eqref{optim:single_kkt}:
\begin{align}
\label{optim:multi_kkt}
    &\tau'=\underset{\substack{v\in\Delta^L\\\sigma_{+,k}, \sigma_{-,k}, \sigma^{[j]}_{+,k}, \sigma^{[j]}_{-,k}\geq 0, \forall j}}{\mathrm{max}}~g({\widebar{x}}^*, {\widebar{u}}^*)^Tv \\
    &s.t.~~~\nabla_{u_k}C^*_{1:n_p-1}+\nabla_{u_k}g({\widebar{x}}^*, {\widebar{u}}^*)^Tv+\sigma_{+,k}+\sigma_{-,k} = 0\nonumber\\
    &~~~~~~~\nabla_{u^{[j]}_k}g({\widebar{x}}^*, {\widebar{u}}^*)^Tv+\sigma^{[j]}_{+,k}+\sigma^{[j]}_{-,k} = 0\nonumber\\
    &~~~~~~~\mathrm{Complementary~Slackness}(\sigma)\nonumber
\end{align}
with $C^*_{1:n_p-1}=C_{1:n_p-1}(x^*, u^*,w_{d-1})$. $\sigma_{+,k}, \sigma_{-,k}, \sigma^{[j]}_{+,k}, $ $\sigma^{[j]}_{-,k}$ denote the multipliers for component-wise action boundaries. Notably, at each stage, only the optimal prepare trajectory $\{x^*_k, u^*_k\}_{k=1}^{n_p}$ and the optimal react trajectory $\{x^{*[j]}_k, u^{*[j]}_k\}_{k=n_p+1}^{N}$ for the realized disturbance $w^{[j]}$ are observed. Since solving problem \eqref{optim:multi_kkt} requires optimal state action sequences for \textit{all} disturbance (scenario) branches, one needs to first infer the optimal react trajectories $\{x^{*[j]}_k, u^{*[j]}_k\}_{k=n_p+1}^{N}$ for \textit{unrealized} disturbances.

We assume the cost function $C(x, u, w)$ is computed as a linear combination of $H$ convex features $\{\Phi_i\}_{i=1}^H$ whose weights $\{\alpha_i\}_{i=1}^H$ are unknown, i.e., $C(x,u,w)=\alpha^T\Phi(x,u,w)$. Then, treating $\{u^{*[j]}_k\}_{k=n_p+1}^{N}$ as the solution to an optimal control problem \eqref{optim:react} over the react phase \cite{majumdar_risk-sensitive_2017}, one can collect observed react sequences from all stages and retrieve weight vector $\alpha\in\mathbb{R}^H$ leveraging inverse KKT method \cite{englert_inverse_2017}. 
\begin{align}
\label{optim:react}
    \underset{\substack{u^{[j]}_k,~k\in\{n_p+1,\dots,N\}}}{\mathrm{min}}~C_{n_p+1:N}(x, u, w)\Big\vert_{w=w^{[j]}}
\end{align}
Then, one could solve the forward optimal control problem and collect react sequences for unrealized disturbances. Finally, the half-space constraint on $\mathcal{P}$ produced by a single stage demonstration is given by $g({\widebar{x}}^*, {\widebar{u}}^*)^Tv\leq\tau',~\forall v\in\mathcal{P}$.

\begin{figure*}[t!]
    \centering
    	\includegraphics[width=0.95\textwidth]{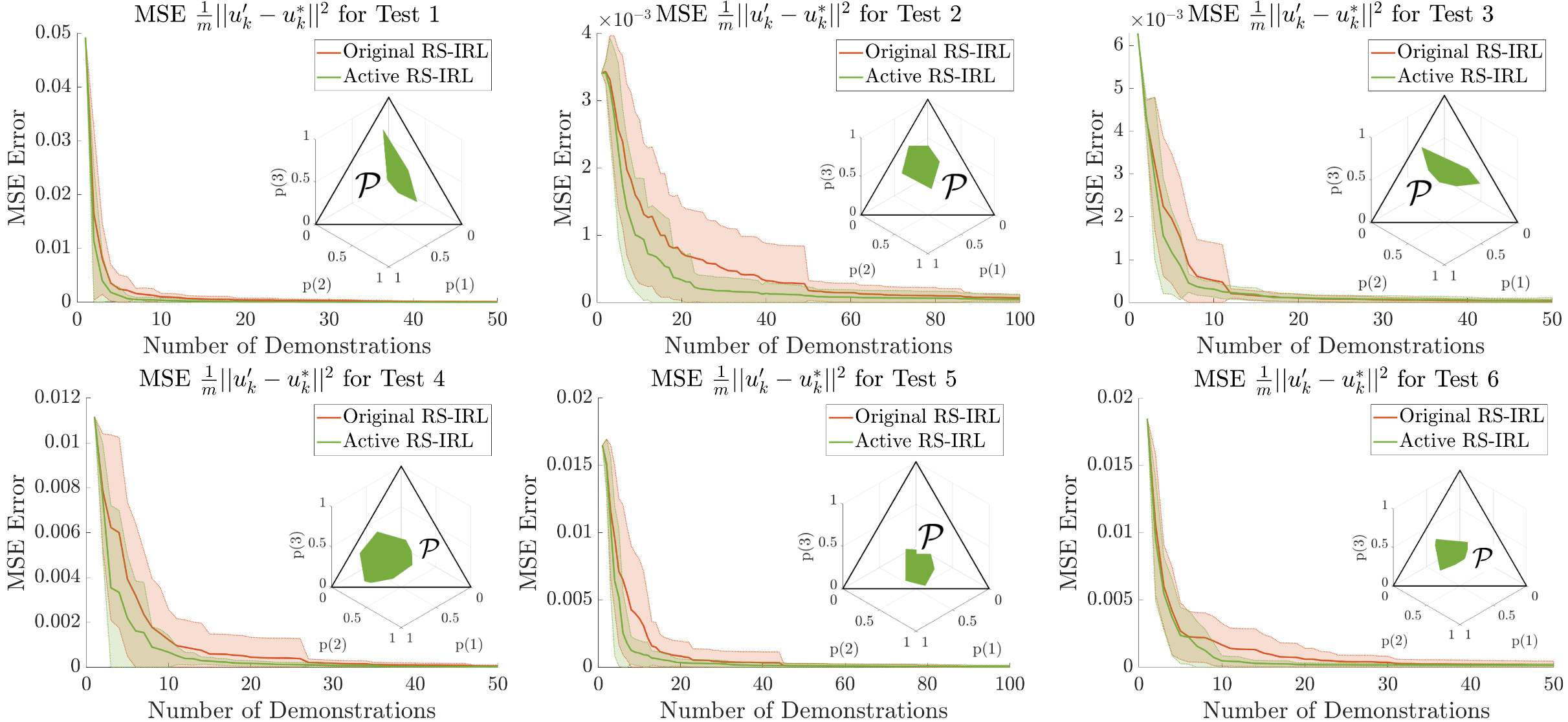}
    \caption{Benchmark results of original single-step RS-IRL algorithm (red) and RS-IRL with active learning (green) on randomly generated dynamical systems and experts. The ground truth risk envelope $\mathcal{P}$ for each test case is also shown. MSE error on action prediction with envelope approximation $\mathcal{P}_d$ at each step is plotted. Shaded area indicates the step-wise standard deviation of prediction error across $30$ training episodes.}
    \label{fig:single_benchmark}
\end{figure*}

\subsection{Adaptive Disturbance Sampling for Multi-Step RS-IRL}
\label{sec:multi_active}

In multi-step setting, since the disturbance is sampled every $N$ steps, the robot evaluates each disturbance $w^{[j]}$ at step $n_p$ according to \eqref{def:dist_preference}, and samples each disturbance exponentially likely with respect to its preference $U_d(j)$. We assume that the disturbance is sampled according to a predefined distribution for $D_0$ initial ``ramp-up'' stages, from which we can collect react sequences for cost function recovery via inverse KKT method. 

In order to extend disturbance evaluation \eqref{def:dist_preference} to multi-step case, we adapt the construction of predicted refinement directions $\prefdir{}\defeq\nrefdir{}({\widebar{x}}', {\widebar{u}}')$. After observing expert action at step $n_p$ of stage $d$, we first predict the expert's react sequences $\{u'^{[j]}_k\}_{k=n_p+1}^{N}$ by solving \eqref{optim:react} for each disturbance. To make the algorithm tractable, we predict the expert react sequences using a finite set of action sequences, i.e., $\mathcal{U}_r$, based on observed expert actions in the react phase. The result is a tail cost vector $g(\widebar{x}, \widebar{u})$ which can be regarded as a priori before observing expert react sequence. To sample $\nrefdir{d}(\widebar{x}',\widebar{u}')$ for disturbance $w_d=w^{[j]}$, we first sample a react sequence from $\mathcal{U}_r$ and compute $g(\widebar{x}', \widebar{u}')(j)$. Then, we compute $\nrefdir{d}(\widebar{x}',\widebar{u}')$ by replacing the $j^{th}$ component of $g(\widebar{x}, \widebar{u})$ by $g(\widebar{x}', \widebar{u}')(j)$ and then projecting the altered cost vector onto probability simplex $\Delta^L$.
Finally, $\nrefdir{d}(\widebar{x}',\widebar{u}')$ is evaluated against previous refinement directions $\{\nrefdir{k}({\widebar{x}}^*,{\widebar{u}}^*)\}_{k\in\mathcal{K}}$ to determine whether it is likely to capture unrevealed envelope boundaries.

Conceptually, at each stage, only one component of the cost vector $g({\widebar{x}}^*, {\widebar{u}}^*)$ will receive information provided by expert demonstration. Other cost realizations are approximated by a deterministic optimal control problem. Thus, the performance of multi-step RS-IRL model heavily depends on the quality of cost function recovery. We leave the study with alternative cost functions (e.g., deep cost functions) as future work and focus on adaptive disturbance sampling under current cost function setting.

\section{Experiments and Analysis}
\label{sec:experiments}

\subsection{Single-Step RS-IRL with Linear-Quadratic Systems}
\label{sec:single_exp}

In this section, we validate the single-step RS-IRL algorithm with adaptive disturbance sampling. We apply our approach to randomly generated systems and experts similar to those in \cite{singh_risk-sensitive_2018}. The state dimension $n$ is $10$. The action dimension $m$ is $5$. We set $L=3$ for visualization purpose. The dynamical system is given by:
\begin{equation}
f(x_k, u_k, w_k) = A(w_k)x_k+B(w_k)u_k
\end{equation}
Each realization $A(w^{[j]})\in\mathbb{R}^{n\times n}$ and $B(w^{[j]})\in\mathbb{R}^{n\times m}$ is generated by independently sampling matrix elements from standard normal distribution $\mathcal{N}(0,1)$. The cost function $C$ is defined as $C(x_k, u_k, w_k) = {u_k}^TRu_k+{x_{k+1}}^TQx_{k+1}$ where $R\succ 0$ and $Q\succeq 0$. $x_{k+1}=f(x_k,u_k,w_k)$. We take $R$ as identity matrix $I_m$. $Q$ is a random positive semi-definite matrix. The initial state $x_0$ is sampled from standard multivariate normal $\mathcal{N}(\mathbf{0},I_n)$ and stays fixed with each system. The expert risk envelope $\mathcal{P}$ is generated by taking the convex hull of $20$ randomly generated points on the probability simplex $\Delta^L$.

We randomly generate six different system-expert pairs and apply single-step RS-IRL algorithm both in original version and with adaptive disturbance sampling. Under each system-expert setup, both methods are repeated for $30$ independent episodes with length of $100$ steps. For testing purpose, we collect $20$ independent episodes, each with $10$ steps. We implement the algorithm in MATLAB and solve the optimization problems using MOSEK \cite{mosek} (with YALMIP interface \cite{Lofberg2004}).

For both original and active RS-IRL algorithms, the approximated $\mathcal{P}_d$ at each step is tested with the $200$ testing state-action pairs $\{x^*_k,u^*_k\}_{k=1}^{200}$. The prediction $u'_k$ is made by solving problem \eqref{optim:single_alt} with respect to $\mathcal{P}_d$ given state $x^*_k$ and evaluated in terms of mean square error (MSE) $\frac{1}{m}||u'_k-u^*_k||^2$. The testing error is averaged over all $30$ episodes per setup. See Fig.~\ref{fig:single_benchmark} for a comparison between original single-step RS-IRL algorithm and our active learning method. Although the original version is also capable of converging rapidly, slow convergence constantly occur due to redundant demonstrations in particular episodes. This yields a high variance in the learning process in terms of error drop. With adaptive disturbance sampling, however, informative demonstrations are queried by the RS-IRL learner with higher probability at earlier steps, which yields progressive envelope refinements.  Thus, we have shown that RS-IRL algorithm with active learning modifications converges faster with lower variance.

\subsection{Multi-Step RS-IRL with Realistic Driving Task}

We apply multi-step RS-IRL algorithm with active learning on a single-lane car-following task in CARLA \cite{Dosovitskiy17}, an open-source driving simulator. The human participant controls a car following another car controlled by an RS-IRL learner. Human control is done via Logitech G920 controller. The state vector $x$ encodes the longitudinal location and velocity of both cars, i.e., $x\defeq[x_f, v_f, x_l, v_l]^T$. Human action $u$ is acceleration of the car. The vehicle dynamics is given by:
\begin{equation}
    \begin{bmatrix} 
    x_{k+1} \\
    v_{k+1} 
    \end{bmatrix} = 
    \begin{bmatrix} 
    1 & \Delta_t \\
    0 & 1 
    \end{bmatrix}
    \begin{bmatrix} 
    x_k \\
    v_k 
    \end{bmatrix}+
    \begin{bmatrix} 
    \frac{1}{2}\Delta^2_t\\
    \Delta_t
    \end{bmatrix}u_k
\end{equation}
for both leading and following car. $\Delta_t=0.1$ s refers to the discrete time step. We collect data form the simulator at $60$ Hz and downsample to $10$ Hz with interpolation.

In this task, the behavior of the leading car is regarded as the source of disturbance to human participant and is sampled every $4$ seconds. We define $L=3$ behaviors from which the leading car may choose to execute: keeping speed, acceleration, and deceleration. Each human participant is told to follow the leading car without colliding. We inform the participants that the leading car will choose and execute one of three maneuvers every $4$ seconds. In the first $D_0 = 40$ ramp-up stages, we sample the leading car behavior according to a fixed $p=[0.3,0.4,0.4]$ to let the human participant establish a mental model of the leading car behavior. Since the true sampling probability is unknown to human participants, we consider the participants ambiguous about the disturbance distribution. We continue the simulation for another $40$ stages with disturbance sampled according to $p$.

\begin{figure}
    \centering
    	\includegraphics[width=0.85\columnwidth]{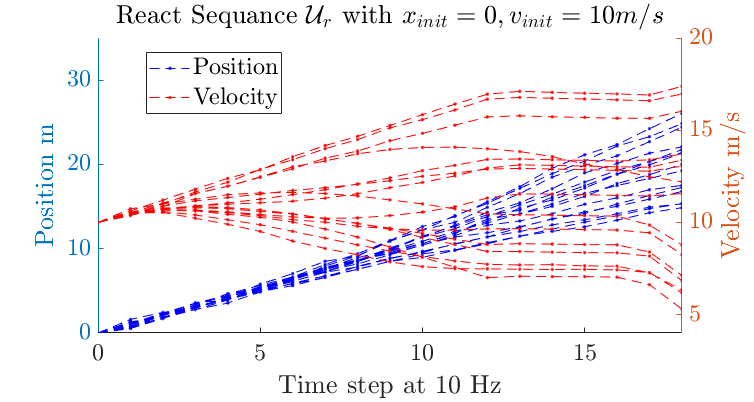}
    \caption{Predicted position and velocity of clustered expert react sequences.}
    \label{fig:react}
\end{figure}

\begin{figure}
    \centering
    	\includegraphics[width=\columnwidth]{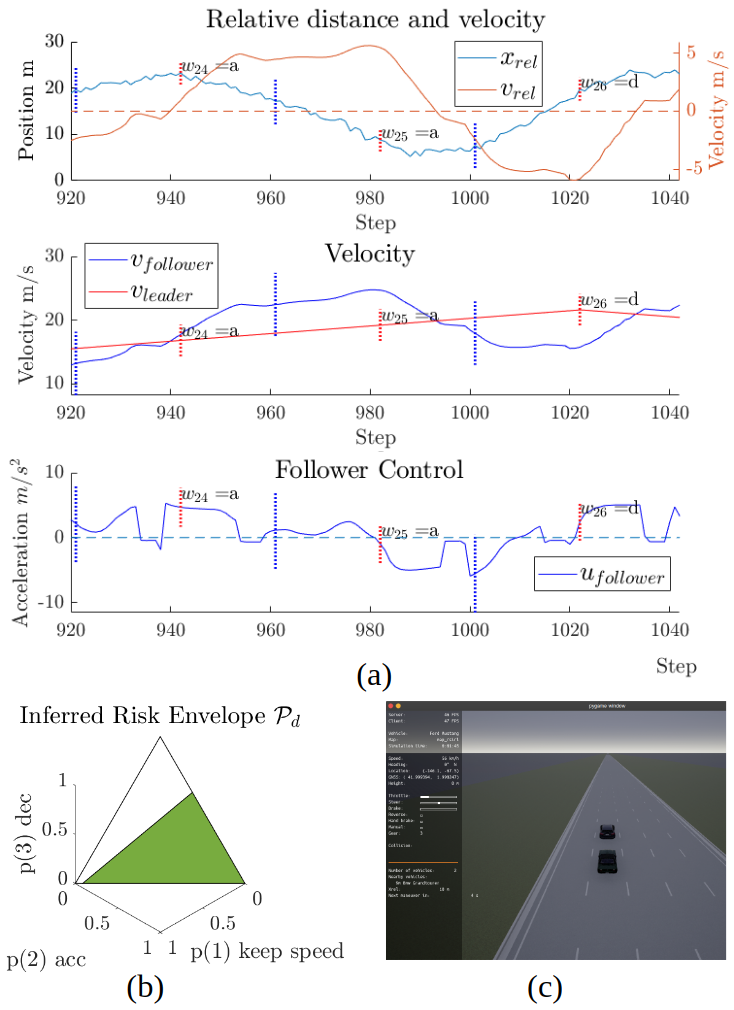}
    \caption{(a) shows collected data in three stages. Stage 25 (middle) produces the dominating half-space constraint for the inferred risk envelope (b) of a risk-seeking participant. Human planning and disturbance sampling periods are separated by blue and red vertical lines respectively. (c) shows the simulation interface. The relative distance and velocity between two cars are shown on the left panel during the simulation. 
    }
    \label{fig:simulation_result}
\end{figure}

For comparison between RS-IRL algorithm with and without active learning, we need separate episodes during which disturbances are sampled differently. In practice, participants usually cannot act fully according to their original intentions due to unfamiliarity with the platform (as opposed to driving a real car) and tend to change their driving style between continuous simulations to ``do better''. This makes repeated experiments very hard. However, most participants are able to maintain the same assessment of disturbances during continuous simulation. Therefore, instead of performing repeated experiment with original and active RS-IRL method, we study a single continuous episode and analyze the disturbances at stages that yield envelope refinements. When analyzing each realized disturbance, we assume that human prepare phase lasts for $2.2$ seconds before observing the new disturbance, i.e., $n_p = 22$ at $10$ Hz. Then, the human participant reacts to the disturbance for $1.8$ seconds. The quadratic features $\Phi(x_k, u_k, w_k)$ used in cost function recovery are defined similarly to those in \cite{singh_risk-sensitive_2018} as following:
\begin{align}
    \Phi_1 &= \mathbf{1}_{x_{rel}<x_0}[\log(1+e^{(-r_1(x_{rel}-x_0))})-\log(2)]\nonumber\\
    \Phi_2 &= \mathbf{1}_{x_{rel}>x_0}[\log(1+e^{(r_2(x_{rel}-x_0))})-\log(2)]\nonumber\\
    \Phi_3 &= r_3(u_k-u_{k-1})^2\nonumber\\
    \Phi_4 &= \log(1+e^{r_4|v_{rel}|})-\log(2)
\end{align}
where $x_{rel}$ and $v_{rel}$ are the follower's relative distance and velocity at the next time step with respect to the leader. The first two features keeps the follower from the leader within a critical distance $x_0=7$ m with a high cost when two cars are too close. Feature $3$ avoids rapidly changing controls. Feature $4$ penalizes high relative speeds. The feature parameters are $r_1 = 3,r_2 = 0.5, r_3 = 0.5, r_4 = 1$. We discretize the human expert's react sequences using K-means clustering with $K=15$. See Fig.~\ref{fig:react} for simulated longitudinal trajectories for each react sequence in the discrete set $\mathcal{U}_r$, which covers full acceleration, deceleration, and varying speeds. 
See Fig.~\ref{fig:simulation_result} for an example of inferred risk envelope for an risk-seeking participant. Note that in stage $25$, the participant keeps a high speed even when the two cars are already approaching rapidly for $\sim4$ s. This means the expert assesses a low probability that the leading car will decelerate, otherwise the expert would decelerate in prepare phase to avoid collision. Thus, stage $25$ is representative of the human's risk preferences and yield a dominating constraint on $\mathcal{P}$, which forms the major boundary of the inferred envelope in \ref{fig:simulation_result}(b). Our active learning scheme provides a disturbance sampling probability of $p=[0.3802, 0.4227, 0.1971]$ at stage $25$, which indeed prefers the acceleration maneuver.

\section{Conclusion and Discussion}\label{sec:conclusion}

In this paper, we present an active learning scheme for risk-sensitive inverse reinforcement learning (RS-IRL). Although the original RS-IRL algorithms solve an approximation of the human risk envelope efficiently, they could suffer slow convergence due to redundant demonstrations. In this research, we develop an adaptive disturbance sampling scheme to enable an RS-IRL learner to actively query potentially informative demonstrations. We evaluate a disturbance realization efficiently by computing future refinement directions on the probability simplex without actually solving any optimization problem. We verify through extensive experiments that our method accelerates the convergence of single-step RS-IRL with lower variance across repeated episodes. We also generalize our method to multi-step setting and apply to a simulated driving task. We have verified that the disturbances leading to successful risk envelope refinements indeed appear preferable in an active learning perspective.

One assumption made in this work is that human's ambiguous assessment of disturbance distribution is fixed. In fact, humans can adapt to observed disturbances and adjust their risk assessments. The constant envelope assumption is not uniquely required by our active learning method, but rather embedded in the original formulation of RS-IRL models. Viewing the original RS-IRL algorithms as capturing \textit{existed} risk preferences, we focus on the active learning under such setting. However, we believe that our methodology can be modulated into more general RS-IRL algorithms in the future; as long as any acquired knowledge on expert risk preferences is still considered valid, our active learning scheme would provide insight on querying informative demonstrations.

\section*{Acknowledgment}
This project is funded in part by Carnegie Mellon University’s Mobility21 National University Transportation Center, which is sponsored by the US Department of Transportation.









\bibliographystyle{IEEEtran}
\bibliography{ref}

\end{document}